%% file: main.tex
\documentclass[10pt,twocolumn,letterpaper]{article}

\usepackage[pagenumbers]{cvpr} 

\input{preamble}
\definecolor{cvprblue}{rgb}{0.21,0.49,0.74}
\usepackage[pagebackref,breaklinks,colorlinks,allcolors=cvprblue]{hyperref}

\usepackage{booktabs}
\usepackage{amsmath}
\usepackage{multirow}
\usepackage{pifont}
\usepackage[table]{xcolor}

\def\confName{CVPR}
\def\confYear{2027}

\title{SCRA-CLIP: Spherical–Contextual Representation Adaptation for Zero-Shot Anomaly Detection}

\author{
Peng Chen$^{1}$ \quad Chao Huang$^{1}$ \quad Wenqiang Wang$^{1}$ \quad
Mingbo Yang$^{1}$ \quad Yannan Chen$^{1}$ \\
Yunkang Cao$^{2}$ \quad Wei Wang$^{1}$ \quad
Li Shen$^{1}$ \quad Fangjun Huang$^{1}$ \\
\\
$^{1}$Shenzhen Campus of Sun Yat-sen University \quad
$^{2}$Hunan University
}

\begin{document}
\maketitle
\input{sec/0_abstract}    
\input{sec/1_intro}
\input{sec/2_formatting}

\input{sec/3_finalcopy}
{
    \small
    \bibliographystyle{ieeenat_fullname}
    \bibliography{main}
}


\end{document}

%% file: sec/0_abstract.tex
\begin{abstract}
Recent advances in vision-language models have facilitated zero-shot anomaly detection (ZSAD) through transferable visual-semantic representations. However, detecting fine-grained anomalies remains challenging due to inadequate modeling of diverse anomaly patterns and their spatial context. To address these challenges, we propose SCRA-CLIP, a Spherical–Contextual Representation Adaptation framework that jointly enhances anomaly-aware textual semantics and context-sensitive visual representations. Specifically, we introduce Spherical Prototype-Conditioned Prompting (SPCP), which employs von Mises--Fisher mixtures to model the directional distributions of normal and abnormal patch features. These state-specific prototypes are incorporated into textual contexts to guide the learning of more discriminative textual representations. Furthermore, we develop Context-Deviation Guided Visual Adaptation (CGVA), which captures local feature deviations relative to multi-scale neighborhood contexts and refines patch representations through adaptive gating. By integrating distribution-aware semantic modeling with context-sensitive visual adaptation, SCRA-CLIP facilitates more discriminative cross-modal matching for fine-grained anomaly detection and localization. Extensive experiments on 11 industrial and medical benchmarks demonstrate the effectiveness and generalization capability of SCRA-CLIP across diverse unseen categories and domains. The code will be publicly available upon publication.
\end{abstract}

%% file: sec/1_intro.tex
\section{Introduction}

\begin{figure}
    \centering
    \includegraphics[width=\linewidth]{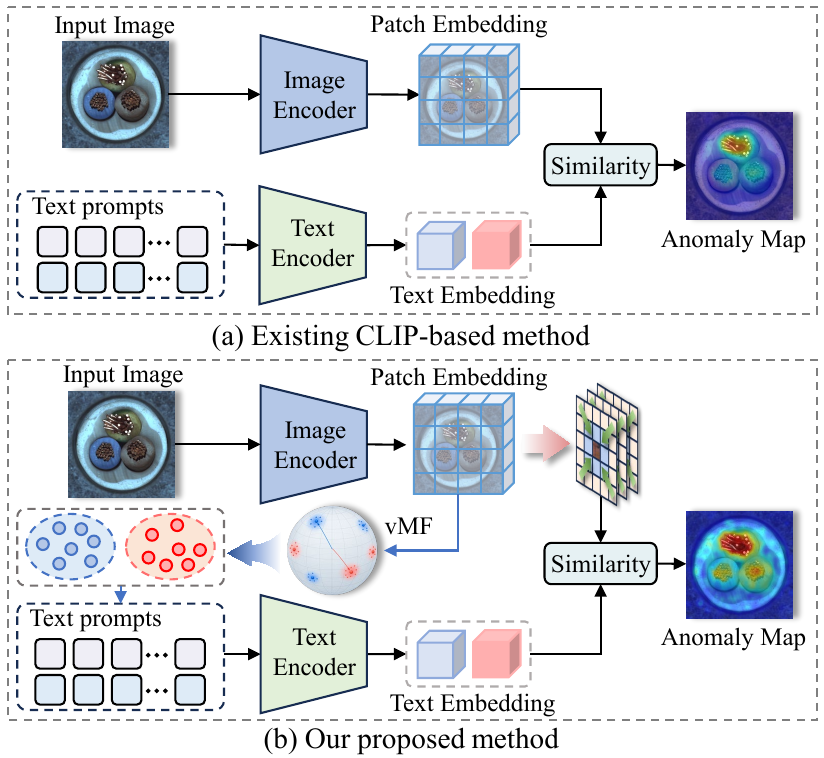}
    \caption{Comparison of existing CLIP-based ZSAD methods and SCRA-CLIP. (a) Existing methods rely on image-text matching with limited local semantic and contextual modeling. (b) SCRA-CLIP enhances anomaly discrimination through spherical prototype modeling and context-guided visual adaptation.}
    \label{fig:intro}
\end{figure}
Anomaly detection aims to identify anomalous images and localize
the corresponding regions, with applications in industrial inspection
and medical image analysis~\cite{qu2025bayesian,huang2026noise,zhang2026logico}.
Conventional methods typically rely on category-specific training data,
limiting their generalization to unseen categories~\cite{roth2022towards,chen2026dyc,gu2024filo}.
Zero-shot anomaly detection (ZSAD) addresses this limitation by detecting
and localizing anomalies without requiring task-specific training data
from target categories. Pretrained vision--language models (VLMs),
particularly CLIP~\cite{radford2021learning}, have facilitated ZSAD
by leveraging transferable visual-semantic representations to match
visual features with textual descriptions of normal and abnormal states~\cite{ma2025aa,shao2026promptmoe}.
Despite this progress, detecting subtle anomalies remains challenging,
as anomalous patterns often exhibit only minor differences from normal
appearances, making their discriminative local cues difficult to capture through global image representations alone.

Recent studies have adapted pretrained VLMs to ZSAD through prompt
learning and visual feature adaptation~\cite{park2026moeclip}.
WinCLIP~\cite{jeong2023winclip} combines compositional text prompts
with multi-scale window matching, whereas
AnomalyCLIP~\cite{zhou2024anomalyclip} learns object-agnostic prompts
to characterize normal and abnormal states. AdaCLIP~\cite{cao2024adaclip}
and VCP-CLIP~\cite{qu2024vcp} incorporate visual cues into textual
prompts. Bayes-PFL~\cite{qu2025bayesian} explores probabilistic
prompt modeling using normalizing flows, and AA-CLIP~\cite{ma2025aa}
introduces anomaly-aware textual anchors with residual adapters.
AdaptCLIP~\cite{gao2026adaptclip} separately trains lightweight visual
and textual adapters with frozen CLIP encoders, while FB-CLIP~\cite{hu2026fb} disentangles foreground and background features.

Despite these advances, two challenges remain in fine-grained ZSAD.
First, normal and abnormal patches exhibit diverse appearance
patterns. Normal patches vary in material, texture, and part
geometry, while abnormal patches exhibit diverse defect types
and appearances. Although existing methods leverage visual
information to adapt textual prompts, explicitly modeling the
multimodal distributions of local patterns within each state
remains underexplored. Since $\ell_2$-normalized patch embeddings
lie on a unit hypersphere, different appearance patterns can
correspond to distinct directions. Modeling multiple directional
modes therefore provides structured visual cues for image-conditioned
textual prompting. Second, anomaly discrimination depends not
only on individual patch appearances but also on their spatial
relationships with surrounding regions. For instance, a patch
in a repetitive texture may appear normal in isolation while
disrupting the surrounding texture pattern. Although visual
encoders may implicitly capture contextual information, conventional
patch-wise matching does not explicitly characterize deviations
from multi-scale neighborhood context. These observations
motivate jointly modeling state-specific appearance distributions
and context-dependent local deviations.

To address these challenges, we propose SCRA-CLIP, a
Spherical--Contextual Representation Adaptation framework for
zero-shot anomaly detection. Figure~\ref{fig:intro} contrasts
SCRA-CLIP with a representative CLIP-based ZSAD pipeline,
highlighting the complementary adaptations to textual prompts
and visual patch features before image--text matching.
Specifically, we introduce Spherical Prototype-Conditioned
Prompting (SPCP), which employs separate von Mises--Fisher (vMF)
mixtures to model the directional distributions of normal and
abnormal patch embeddings and uses cross-attention to incorporate
the learned state-specific prototypes into image-conditioned
textual prompts. In parallel, we develop Context-Deviation Guided
Visual Adaptation (CGVA), which explicitly models local patch
deviations from multi-scale neighborhood context and adaptively
refines patch representations through constrained residual updates.
By jointly enhancing textual semantics and contextual visual
representations, SCRA-CLIP facilitates more discriminative
cross-modal matching for fine-grained anomaly detection and
localization. The main contributions of this work are summarized as follows:

\begin{itemize}
    \item We propose SCRA-CLIP, a unified framework that jointly adapts textual prompts and local visual features for image-level anomaly detection and pixel-level localization in unseen categories.

    \item We introduce SPCP, which employs vMF mixtures to model the directional distributions of normal and abnormal patch features and incorporates state-specific prototypes into textual contexts through cross-attention.

    \item We develop CGVA, which explicitly models local feature deviations relative to multi-scale neighborhood context to improve fine-grained anomaly localization.

    \item Extensive experiments on 11 industrial and medical benchmarks demonstrate that SCRA-CLIP achieves outstanding ZSAD performance.
\end{itemize}

%% file: sec/2_formatting.tex
\section{Related Work}

\subsection{Zero-shot Anomaly Detection}

Zero-shot anomaly detection (ZSAD) aims to identify and localize anomalies in unseen categories without target-category training data~\cite{xu2025towards,li2026musc}. Leveraging the transferable visual-semantic representations of vision-language models such as CLIP, recent methods have achieved promising cross-category generalization~\cite{fang2025af,he2025rareclip,sadikaj2025multiads}. WinCLIP~\cite{jeong2023winclip} employs compositional text prompts and multi-scale window matching, while AnomalyCLIP~\cite{zhou2024anomalyclip} learns object-agnostic prompts to capture category-independent anomaly semantics. AA-CLIP~\cite{ma2025aa} introduces anomaly-aware textual anchors and residual adapters for fine-grained cross-modal alignment. VisualAD~\cite{hou2026visualad}
explores language-free anomaly detection using a vision transformer. Despite these advances, effectively capturing context-dependent local variations while preserving the generalization of pretrained representations remains challenging. Our SCRA-CLIP addresses this limitation through distribution-aware
prompt modeling and context-sensitive visual adaptation.

\begin{figure*}
    \centering
    \includegraphics[width=\linewidth]{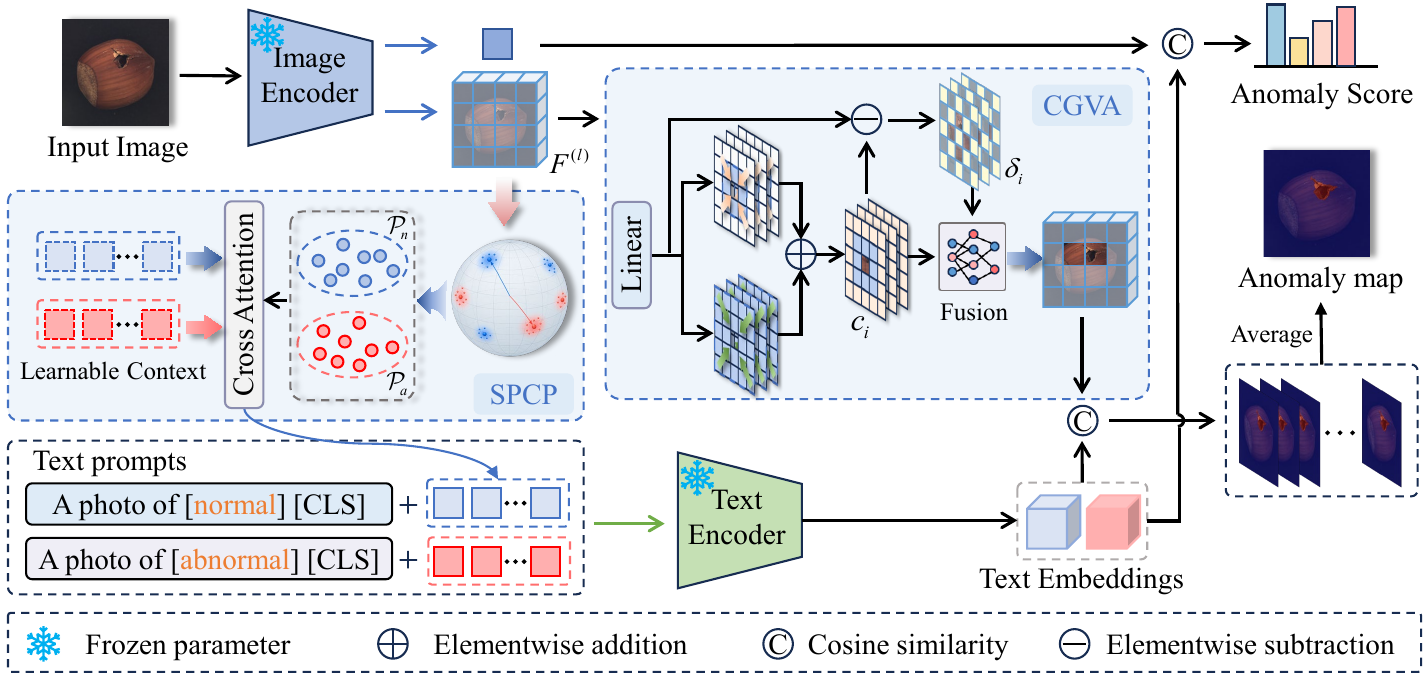}
    \caption{Overall architecture of SCRA-CLIP. SPCP learns state-specific spherical prototypes to enhance anomaly-aware textual representations, while CGVA refines local visual features through multi-scale context-deviation modeling and constrained residual adaptation. Their complementary adaptation improves cross-modal matching for zero-shot anomaly detection and localization.}
    \label{fig:method}
\end{figure*}

\subsection{Prompt Learning}

Prompt learning provides a flexible strategy for adapting
pretrained vision--language models to downstream tasks~\cite{wei2026exploring,ma2025aa,salehi2025crane}.
CoOp~\cite{zhou2022learning} introduces learnable textual prompts,
while CoCoOp~\cite{zhou2022conditional} incorporates
image-conditioned prompts to improve generalization. In ZSAD,
AdaCLIP~\cite{cao2024adaclip} combines static and dynamic prompts,
whereas VCP-CLIP~\cite{qu2024vcp} integrates global and local
visual context into textual representations. Bayes-PFL
~\cite{qu2025bayesian} models probabilistic prompt distributions
using normalizing flows. AdaptCLIP~\cite{gao2026adaptclip}
further explores cross-modal adaptation through lightweight
visual and textual adapters. Our SPCP explicitly models the
directional distributions of local normal and abnormal patch
embeddings using von Mises--Fisher mixtures and incorporates
state-specific prototypes into textual prompts through
cross-attention.

%% file: sec/3_finalcopy.tex

\section{Methodology}

\subsection{Problem Definition}

Zero-shot anomaly detection (ZSAD) aims to detect and localize anomalies in unseen categories without target-category training data. Given an auxiliary training dataset $\mathcal{D}_s=\{(x_i,y_i,M_i)\}_{i=1}^{N_s}$, where $x_i\in\mathbb{R}^{H\times W\times 3}$ is an input image, $y_i\in\{0,1\}$ denotes its anomaly label, and $M_i\in\{0,1\}^{H\times W}$ is the corresponding anomaly mask, the model learns transferable representations from seen categories $\mathcal{C}_s$. During inference, the trained model is directly evaluated on unseen categories $\mathcal{C}_u$, where $\mathcal{C}_s\cap\mathcal{C}_u=\emptyset$, without further adaptation. For each test image $x$, the model predicts an image-level anomaly score $s(x)\in[0,1]$ and a pixel-level anomaly map $A(x)\in[0,1]^{H\times W}$ for anomaly detection and localization, respectively.

\subsection{Overview}

Figure~\ref{fig:method} illustrates the overall architecture of SCRA-CLIP. Given an input image, the frozen visual encoder extracts a global image feature and multi-layer patch features. Spherical
Prototype-Conditioned Prompting (SPCP) models the directional
distributions of normal and abnormal patch features using
state-specific von Mises--Fisher mixtures. The learned prototypes
are incorporated into learnable textual contexts through
cross-attention to generate discriminative text embeddings. Context-Deviation Guided Visual Adaptation (CGVA) explicitly models local feature deviations relative to multi-scale neighborhood context and refines patch representations through adaptive gating and residual updates. By integrating distribution-aware textual adaptation with context-sensitive visual refinement, SCRA-CLIP enables more discriminative cross-modal matching for anomaly detection and localization in unseen categories.

\subsection{Spherical Prototype-Conditioned Prompting}

Fixed textual representations can struggle to capture the diversity
of local normal and abnormal patterns across images and object
categories. We propose Spherical Prototype-Conditioned Prompting
(SPCP), which models the directional distributions of normalized
patch embeddings using state-specific von Mises--Fisher (vMF) mixtures.
Aggregated patch features from the input image condition the
cross-attention queries, which attend to the corresponding prototype
banks to generate image-conditioned normal and abnormal text embeddings.

\noindent\textbf{Spherical Prototype Modeling.}
Given the patch features extracted from the $l$-th visual layer, denoted as $F^{(l)}\in\mathbb{R}^{N\times D_v}$, where $N$ and $D_v$ denote the number of patches and feature dimension, respectively, we project them into a compact embedding space followed by $\ell_2$ normalization:
\begin{equation}
Z^{(l)}=\mathrm{Norm}\left(\phi_l(F^{(l)})\right)
\in\mathbb{R}^{N\times d},
\end{equation}
where $\phi_l(\cdot)$ denotes a learnable layer-specific projection and $\mathrm{Norm}(\cdot)$ represents $\ell_2$ normalization. The resulting embeddings lie on a unit hypersphere, making von Mises--Fisher (vMF) mixtures suitable for modeling their directional distributions.
Specifically, we introduce two learnable prototype banks for normal and abnormal states at each selected layer, denoted as $\mathcal{P}_{s}^{(l)}=\{\mu_{s,k}^{(l)}\}_{k=1}^{K}$, where $s\in\{n,a\}$ and $K$ denotes the number of prototypes per state. Each prototype is normalized as $\hat{\mu}_{s,k}^{(l)}=\mu_{s,k}^{(l)}/\|\mu_{s,k}^{(l)}\|_2$. Let $\hat{z}_i^{(l)}$ denote the normalized embedding of patch $i$,
corresponding to the $i$-th row of $Z^{(l)}$. The weighted component
log-score for state $s$ is given by:
\begin{equation}
e_{i,s,k}^{(l)}
=\kappa(\hat{z}_i^{(l)})^{\top}\hat{\mu}_{s,k}^{(l)}
-\log K,
\end{equation}
where $\kappa$ is fixed and the mixture weights are uniform. The state-specific log-likelihood score, up to the shared log-normalization term, is computed as:
\begin{equation}
S_{i,s}^{(l)}
=\log\sum_{k=1}^{K}\exp(e_{i,s,k}^{(l)}).
\end{equation}

\noindent\textbf{Prototype Learning Objectives.}
To learn state-discriminative prototypes, we introduce a vMF density-fitting loss and a discriminative classification loss. Patch-level labels are obtained by downsampling the ground-truth masks to the feature resolution. Let $\mathcal{I}_s$ denote the set of patch indices belonging to state $s$ across the current mini-batch, and $\mathcal{S}\subseteq\{n,a\}$ denote the set of states present in the mini-batch. The vMF density-fitting loss maximizes the state-specific log-likelihood scores of patch embeddings:
\begin{equation}
\mathcal{L}_{\mathrm{vmf}}
=-\frac{1}{|\mathcal{L}|}
\sum_{l\in\mathcal{L}}
\frac{1}{|\mathcal{S}|}
\sum_{s\in\mathcal{S}}
\frac{1}{|\mathcal{I}_s|}
\sum_{i\in\mathcal{I}_s}
S_{i,s}^{(l)},
\end{equation}
where $\mathcal{L}$ denotes the set of selected visual layers. To further enhance discrimination between normal and abnormal states, we employ a cross-entropy loss over the state-specific mixture scores:
\begin{equation}
\begin{aligned}
\mathcal{L}_{\mathrm{disc}}
=&\frac{1}{|\mathcal{L}|}
\sum_{l\in\mathcal{L}}
\frac{1}{|\mathcal{S}|}
\sum_{s\in\mathcal{S}}
\frac{1}{|\mathcal{I}_s|}
\sum_{i\in\mathcal{I}_s}
\ell_{i,s}^{(l)},\\
\ell_{i,s}^{(l)}
=&-\log
\frac{\exp(S_{i,s}^{(l)})}
{\sum_{s'\in\{n,a\}}\exp(S_{i,s'}^{(l)})}.
\end{aligned}
\end{equation}
The vMF density-fitting loss encourages patch embeddings to align with their corresponding state-specific prototype distributions, while the discriminative loss enhances the separability of normal and abnormal state scores. These objectives facilitate the learning of state-discriminative prototypes, which provide structured visual guidance for subsequent prototype-conditioned prompt generation.

\noindent\textbf{Prototype-Conditioned Prompt Generation.}
For each state $s\in\{n,a\}$, we stack its prototypes across the
selected visual layers and project them into the text token
embedding space to obtain $B_s\in\mathbb{R}^{|\mathcal{L}|K\times D_t}$,
where $D_t$ is the text token embedding dimension. Let
$C_s^0\in\mathbb{R}^{M\times D_t}$ denote the $M$ learnable base
context tokens for state $s$.

To form image-conditioned queries, we aggregate the patch features
$F^{(l)}$ from a selected visual layer $l\in\mathcal{L}$ and add
the resulting vector to each base context token. We project $B_s$
into keys and values using state-specific projections and compute
the cross-attention output as:
\begin{equation}
H_s=\operatorname{Attn}\bigl(
    C_s^0+\mathcal{G}_p(F^{(l)}), B_sW_s^k, B_sW_s^v
\bigr),
\end{equation}
where $\mathcal{G}_p(\cdot)$ averages the $N$ patch tokens and maps
the pooled feature into the text token embedding space via a
learnable linear projection. The learnable matrices $W_s^k$ and
$W_s^v$ are the key and value projections for state $s$,
respectively. We incorporate the aggregated prototype
features into the base context through a residual update:
\begin{equation}
C_s=C_s^0+\gamma_s\operatorname{LN}(H_s),
\end{equation}
where $\gamma_s$ is a learnable state-specific scalar that scales
the prototype update and $\operatorname{LN}(\cdot)$ denotes layer
normalization. We insert $C_s$ into the corresponding prompt template
and encode the resulting prompt with the frozen text encoder to
obtain the image-conditioned text embedding $t_s$ for cross-modal
matching.

\subsection{Context-Deviation Guided Visual Adaptation}

Fine-grained anomaly discrimination relies not only on discriminative patch representations but also on their local spatial context. However, patch-wise representations may fail to capture subtle deviations from neighboring regions, while unconstrained feature adaptation can introduce undesirable perturbations to visual representations. To address these limitations, we propose Context-Deviation Guided Visual Adaptation (CGVA), which explicitly captures context-relative feature deviations at multiple spatial scales and employs constrained residual updates to refine local visual representations.

\noindent\textbf{Multi-Scale Context-Deviation Modeling.}
Given the patch features $F^{(l)}$ extracted from the $l$-th visual layer, we first project them into a shared embedding space:
\begin{equation}
\widetilde{F}^{(l)}=\psi_l(F^{(l)})
\in\mathbb{R}^{N\times D},
\end{equation}
where $\psi_l(\cdot)$ denotes a learnable layer-specific projection and $D$ is the feature dimension. For simplicity, we omit the layer index in the following formulations. Each projected patch feature $\tilde{f}_i$ is normalized as $p_i=\mathrm{Norm}(\tilde{f}_i)$, forming $P=[p_1,\ldots,p_N]^\top\in\mathbb{R}^{N\times D}$.

To characterize local structural variations, we reshape the patch tokens into their spatial grid and aggregate neighborhood information at multiple scales. Specifically, the multi-scale context representation and its corresponding feature deviation are computed as:
\begin{equation}
\begin{aligned}
c_i &= \mathrm{Norm}\left(
[\mathrm{AvgPool}_{3}(P)]_i+
[\mathrm{AvgPool}_{5}(P)]_i
\right),\\
\delta_i &= p_i-c_i,
\end{aligned}
\end{equation}
where $\mathrm{AvgPool}_{k}(\cdot)$ denotes spatial average pooling with a $k\times k$ neighborhood. The context representation $c_i$ captures local neighborhood context, while the deviation $\delta_i$ characterizes the discrepancy between each patch and its local neighborhood. By jointly modeling contextual information and feature deviations, CGVA captures context-relative local variations for subsequent visual adaptation.

\noindent\textbf{Gated Residual Adaptation.}
To adaptively integrate local deviations and neighborhood context, we first project them into a compact feature space and then introduce two separately parameterized gates, jointly conditioned on both representations, to dynamically regulate their contributions:
\begin{equation}
\begin{aligned}
u_i &= W_\delta \delta_i+b_\delta,\quad
v_i=W_c c_i+b_c,\\
g_i^{q}&=2\sigma\left(
W_g^{q}[u_i;v_i]+b_g^{q}
\right),\quad q\in\{\delta,c\},
\end{aligned}
\end{equation}
where $W_\delta$ and $W_c$ are learnable projection matrices, $b_\delta$ and $b_c$ are their corresponding biases, $W_g^q$ and $b_g^q$ denote the learnable weights and biases of the corresponding gates, respectively, $[\cdot;\cdot]$ denotes feature concatenation, and $\sigma(\cdot)$ is the sigmoid function. The gated representations are fused to generate a context-aware residual feature:
\begin{equation}
r_i=W_o\mathrm{GELU}\left(
g_i^\delta\odot u_i+g_i^c\odot v_i
\right)+b_o,
\end{equation}
where $\odot$ denotes element-wise multiplication, and $W_o$ projects the fused representation back to the dimension of the projected patch features.
To prevent excessive directional perturbations in the projected visual feature space, we constrain the residual magnitude and perform a bounded residual-based directional refinement:
\begin{equation}
\bar{r}_i=\frac{r_i}{\max(1,\|r_i\|_2)},\quad
\hat{f}_i=\mathrm{Norm}(p_i+\alpha\bar{r}_i),
\end{equation}
where $\alpha$ controls the residual adaptation strength and satisfies $0\leq\alpha<1$. By bounding the residual magnitude and maintaining unit-normalized patch representations, CGVA enables controlled directional adaptation while limiting excessive perturbations to the visual representations.

\subsection{Anomaly Map and Anomaly Score}

\textbf{Pixel-level Anomaly Map.}
Given the normal and abnormal text embeddings generated by SPCP, denoted as
$T=[t_n,t_a]^\top\in\mathbb{R}^{2\times D}$, we compute pixel-level anomaly
maps by matching them with the CGVA-refined patch features. Specifically,
for each selected visual layer $l$, we stack the refined patch features into
$\hat{F}^{(l)}=[\hat{f}_1^{(l)},\ldots,\hat{f}_N^{(l)}]^\top
\in\mathbb{R}^{N\times D}$. The final anomaly map is obtained by aggregating
the patch-level anomaly probabilities across visual layers:
\begin{equation}
A=
\frac{1}{|\mathcal{L}|}
\sum_{l\in\mathcal{L}}
\operatorname{Up}\left(
\left[
\operatorname{Softmax}\left(
\frac{
\hat{F}^{(l)}\operatorname{Norm}(T)^\top
}{\tau}
\right)
\right]_{:,a}
\right),
\end{equation}
where $[\cdot]_{:,a}$ selects the abnormal probability for all patches,
and $\operatorname{Up}(\cdot)$ denotes spatial reconstruction followed by
bilinear interpolation to the input resolution. The Softmax operation is
applied along the normal and abnormal state dimension. By combining
anomaly-aware textual representations with context-adapted visual features,
the resulting anomaly map integrates complementary predictions across
multiple visual layers for fine-grained anomaly localization.

\noindent\textbf{Image-level Anomaly Score.}
For image-level anomaly detection, we combine the global
image representation with pooled original patch features
to capture complementary global semantic and local visual information.
Specifically, the global image feature $f_g$ and pooled
patch representation $f_p$ are projected into a shared
embedding space and then individually $\ell_2$-normalized.
The fused image representation is computed as:
\begin{equation}
f_{\mathrm{img}}=
\mathrm{Norm}(f_g+f_p).
\end{equation}
The image-level anomaly score is obtained by computing
cosine similarities between the fused image representation
and the normal and abnormal text embeddings:
\begin{equation}
s(x)=
\left[
\mathrm{Softmax}\left(
\frac{
f_{\mathrm{img}}\mathrm{Norm}(T)^\top
}{\tau}
\right)
\right]_a,
\end{equation}
where $\tau$ is the temperature parameter and $[\cdot]_a$
selects the abnormal probability. This fusion combines
global semantics with local visual information for
image-level anomaly discrimination.

\subsection{Loss Function}

We use BCE loss for image-level classification and combine
Dice and Focal losses for pixel-level segmentation.
The pixel-level segmentation loss is formulated as:
\begin{equation}
\mathcal{L}_{\mathrm{px}}
=\frac{1}{|\mathcal{L}|}\sum_{l\in\mathcal{L}}
\left[
\mathrm{Dice}(A^{(l)},M)
+\mathrm{Focal}(A^{(l)},M)
\right],
\end{equation}
where $A^{(l)}$ denotes the predicted anomaly maps from the
$l$-th visual layer, and $M$ represents the corresponding
ground-truth anomaly masks. Finally, the overall training objective is defined as:
\begin{equation}
\begin{aligned}
\mathcal{L}_{\mathrm{total}}
=\mathcal{L}_{\mathrm{px}}+\mathrm{BCE}(\mathbf{s},\mathbf{y})
+\lambda_{\mathrm{vmf}}\mathcal{L}_{\mathrm{vmf}}
+\lambda_{\mathrm{disc}}\mathcal{L}_{\mathrm{disc}},
\end{aligned}
\end{equation}
where $\mathbf{s}$ and $\mathbf{y}$ denote the predicted
image-level anomaly scores and their corresponding
ground-truth labels, respectively. $\lambda_{\mathrm{vmf}}$
and $\lambda_{\mathrm{disc}}$ control the contributions
of the two auxiliary prototype learning losses introduced
in SPCP.

\begin{table*}[t]
\centering
\caption{Comparison of pixel-level AUROC (\%) on industrial and medical anomaly detection datasets. The best and second-best results are highlighted in bold and underlined, respectively.}
\label{tab:pixel_auroc}
\resizebox{\textwidth}{!}{
\begin{tabular}{lccccccccc}
\toprule
\multirow{2}{*}{Method}
& \multirow{2}{*}{Venue}
& \multicolumn{5}{c}{Industrial Defects}
& \multicolumn{3}{c}{Medical Anomalies} \\
\cmidrule(lr){3-7}
\cmidrule(lr){8-10}
& & VisA & MVTec-AD & BTAD & DAGM & DTD-Synthetic
& ISIC & CVC-ClinicDB & Kvasir \\
\midrule
\midrule

CLIP~\cite{radford2021learning} & ICML'21
& 46.6 & 38.4 & 30.6 & 30.6 & 33.9
& 33.1 & 47.5 & 44.6 \\

CoOp~\cite{zhou2022learning} & IJCV'22
& 24.2 & 33.3 & 15.4 & 17.5 & 66.5
& 51.7 & 34.8 & 44.1 \\

WinCLIP~\cite{jeong2023winclip} & CVPR'23
& 79.6 & 85.1 & 72.7 & 87.6 & 83.9
& 83.3 & 51.2 & 69.7 \\

CLIP-AD~\cite{chen2024clip} & IJCAI'24
& 95.0 & 89.8 & 93.1 & 96.5 & 97.1
& 81.6 & 85.8 & 82.5 \\

AnomalyCLIP~\cite{zhou2024anomalyclip} & ICLR'24
& 95.5 & 91.1 & 94.2 & 95.6 & 97.9
& 88.4 & 82.9 & 78.9 \\

AdaCLIP~\cite{cao2024adaclip} & ECCV'24
& 95.1 & 86.8 & 87.7 & 97.0 & 94.1
& 85.4 & 84.4 & 79.4 \\

TPS~\cite{ma2025aligning} & AAAI'25
& 95.5 & 88.5 & 94.2 & 93.8 & 97.9
& 90.2 & 84.8 & 78.8 \\

AA-CLIP~\cite{ma2025aa} & CVPR'25
& 95.5 & \underline{91.9} & \underline{96.1} & 96.3 & 96.4
& \underline{92.8} & 89.5 & \underline{87.6} \\

Bayes-PFL~\cite{qu2025bayesian} & CVPR'25
& 95.6 & \underline{91.9} & 93.9 & \underline{99.3} & 97.8
& 92.2 & \underline{89.6} & 85.4 \\

FAPrompt~\cite{zhu2025fine} & ICCV'25
& \underline{95.9} & 90.6 & 95.6 & 98.2 & \underline{98.3}
& 91.1 & 84.7 & 81.2 \\

AdaptCLIP~\cite{gao2026adaptclip} & AAAI'26
& 95.7 & 90.9 & 93.8 & 94.7 & 97.7
& 91.0 & 83.8 & 86.5 \\

VisualAD~\cite{hou2026visualad} & CVPR'26
& 95.8 & 90.8 & 91.6 & 93.6 & 98.1
& 84.0 & 85.2 & 80.3 \\

\midrule
\rowcolor[HTML]{E6E6E6}
\textbf{SCRA-CLIP} & \textbf{Ours}
& \textbf{96.0} & \textbf{92.3} & \textbf{96.3}
& \textbf{99.5} & \textbf{98.4}
& \textbf{93.2} & \textbf{90.5} & \textbf{87.8} \\

\bottomrule
\end{tabular}
}
\end{table*}

\section{Experiments}
\subsection{Experiment Setup}
\textbf{Datasets.}
We evaluate SCRA-CLIP on 11 publicly available benchmarks, comprising five industrial and six medical datasets. The industrial benchmarks include MVTec-AD~\cite{bergmann2019mvtec}, VisA~\cite{zou2022spot}, BTAD~\cite{mishra2021vt}, DAGM~\cite{wieler2007weakly}, and DTD-Synthetic~\cite{aota2023zero}, covering diverse object categories, textures, and defect patterns. The medical benchmarks include ISIC~\cite{codella2018skin}, CVC-ClinicDB~\cite{bernal2015wm}, Kvasir-SEG~\cite{jha2019kvasir}, HeadCT~\cite{salehi2021multiresolution}, BrainMRI~\cite{kanade2015brain}, and Br35H~\cite{hamada2020br35h}, spanning dermoscopic, endoscopic, CT, and MRI modalities. These benchmarks provide a comprehensive evaluation of SCRA-CLIP across industrial and medical domains.

\noindent\textbf{Evaluation Metrics.}
Following prior work~\cite{zhou2024anomalyclip,qu2025bayesian,gao2026adaptclip},
we adopt the area under the receiver operating characteristic curve (AUROC)
as the evaluation metric. We compute AUROC at the image level (I-AUROC)
to evaluate the model's anomaly detection and at the pixel level (P-AUROC) to assess anomaly localization performance.

\noindent\textbf{Implementation Details.}
We adopt CLIP ViT-L/14-336 with frozen visual and text encoders. All images are resized to $518\times518$, and patch features are extracted from layers 6, 12, 18, and 24. For SPCP, we set the number of prototypes to 16 per state per layer and the vMF concentration parameter to $\kappa=20$. For CGVA, multi-scale context is captured using $3\times3$ and $5\times5$ average pooling, with the residual adaptation strength $\alpha=0.1$. The auxiliary loss weights are set to $\lambda_{\mathrm{vmf}}=0.05$ and $\lambda_{\mathrm{disc}}=0.1$. We train for 20 epochs using Adam with a learning rate of $2\times10^{-4}$ and a batch size of 32. Following prior work~\cite{qu2025bayesian, hou2026visualad}, we train on VisA for evaluation on all other benchmarks and on MVTec-AD for evaluation on VisA dataset. All experiments run on an NVIDIA GeForce RTX 3090 GPU. More details are provided in the Supplementary Material.

\begin{table*}[t]
\centering
\caption{Comparison of image-level AUROC (\%) on industrial and medical anomaly detection datasets. The best and second-best results are highlighted in bold and underlined, respectively.}
\label{tab:image_auroc}
\resizebox{\textwidth}{!}{
\begin{tabular}{lccccccccc}
\toprule
\multirow{2}{*}{Method}
& \multirow{2}{*}{Venue}
& \multicolumn{5}{c}{Industrial Defects}
& \multicolumn{3}{c}{Medical Anomalies} \\
\cmidrule(lr){3-7}
\cmidrule(lr){8-10}
& & VisA & MVTec-AD & BTAD & DAGM & DTD-Synthetic
& HeadCT & BrainMRI & Br35H \\
\midrule
\midrule

CLIP~\cite{radford2021learning} & ICML'21
& 66.4 & 74.1 & 34.5 & 79.6 & 71.6
& 56.5 & 73.9 & 78.4 \\

CoOp~\cite{zhou2022learning} & IJCV'22
& 62.8 & 88.8 & 66.8 & 87.5 & 63.4
& 78.4 & 61.3 & 86.0 \\

WinCLIP~\cite{jeong2023winclip} & CVPR'23
& 78.8 & 91.8 & 68.2 & 91.8 & 93.2
& 83.7 & 86.6 & 80.5 \\

CLIP-AD~\cite{chen2024clip} & IJCAI'24
& 79.8 & 89.8 & 85.8 & 90.8 & 91.5
& 93.8 & 92.8 & 96.0 \\

AnomalyCLIP~\cite{zhou2024anomalyclip} & ICLR'24
& 82.1 & 91.5 & 88.3 & 97.5 & 93.5
& 95.3 & 90.3 & 94.6 \\

AdaCLIP~\cite{cao2024adaclip} & ECCV'24
& 83.0 & 92.0 & 91.6 & 96.5 & 92.8
& 93.3 & 94.8 & 95.7 \\

TPS~\cite{ma2025aligning} & AAAI'25
& 83.3 & 90.1 & 88.1 & 97.0 & 93.4
& 92.3 & 92.4 & 96.2 \\

AA-CLIP~\cite{ma2025aa} & CVPR'25
& 84.6 & 90.5 & \underline{93.8} & 94.3 & 91.4
& 94.8 & 94.1 & \underline{97.8} \\

Bayes-PFL~\cite{qu2025bayesian} & CVPR'25
& \underline{86.8} & 92.2 & 93.2 & \underline{97.7} & 95.1
& \underline{96.5} & 96.2 & \underline{97.8} \\

FAPrompt~\cite{zhu2025fine} & ICCV'25
& 84.6 & 91.9 & 92.2 & \textbf{97.8} & 95.9
& 94.0 & 95.8 & 97.6 \\

AdaptCLIP~\cite{gao2026adaptclip} & AAAI'26
& 84.8 & \textbf{93.5} & 91.0 & 95.2 & 96.0
& 92.6 & 90.1 & 94.8 \\

VisualAD~\cite{hou2026visualad} & CVPR'26
& 84.7 & 92.2 & 93.6 & 97.6 & \underline{97.5}
& 90.3 & \underline{96.7} & 95.5 \\

\midrule
\rowcolor[HTML]{E6E6E6}
\textbf{SCRA-CLIP} & \textbf{Ours}
& \textbf{87.2} & \underline{92.4} & \textbf{94.4}
& \textbf{97.8} & \textbf{98.0}
& \textbf{98.3} & \textbf{96.8} & \textbf{98.4} \\

\bottomrule
\end{tabular}
}
\end{table*}

\subsection{Comparison with State-of-the-Art Methods}

\begin{figure}
    \centering
    \includegraphics[width=\linewidth]{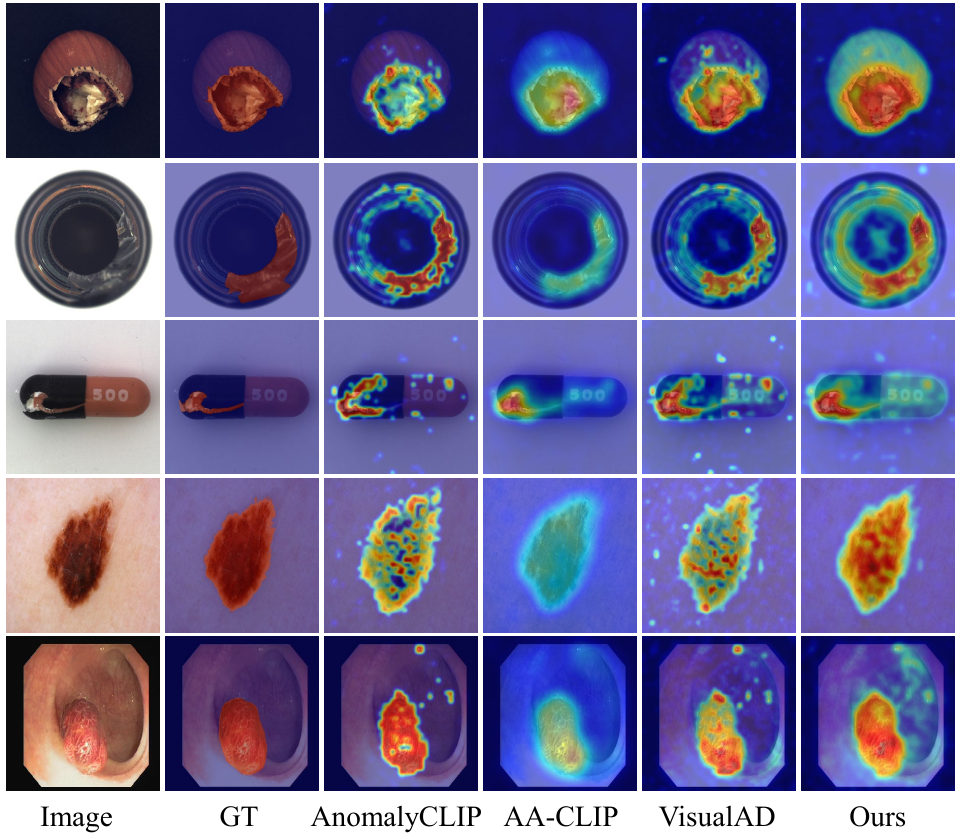}
    \caption{Qualitative comparison of anomaly localization results on industrial and medical datasets.}
    \label{fig:vis}
\end{figure}

We compare SCRA-CLIP with 12 representative methods on industrial and medical anomaly detection benchmarks. More results are provided in the Supplementary Material.

\noindent\textbf{Zero-shot anomaly segmentation.}
As shown in Table~\ref{tab:pixel_auroc}, SCRA-CLIP consistently achieves the highest pixel-level AUROC across all eight benchmarks. It demonstrates strong localization performance on industrial datasets, achieving particularly high AUROCs of 99.5\% on DAGM and 98.4\% on DTD-Synthetic. Notably, SCRA-CLIP also achieves the best results on all three medical benchmarks, with an improvement of 0.9 percentage points over the strongest baseline on CVC-ClinicDB. These results highlight the effectiveness and cross-domain robustness of SCRA-CLIP in fine-grained anomaly localization.

\noindent\textbf{Zero-shot anomaly classification.}
As shown in Table~\ref{tab:image_auroc}, SCRA-CLIP achieves the best or tied-best image-level AUROC on seven of the eight benchmarks, ranking second on MVTec-AD. It achieves the best or tied-best performance on four of the five industrial benchmarks and leads across all three medical benchmarks. Notably, SCRA-CLIP obtains 98.3\% AUROC on HeadCT, surpassing the strongest baseline by 1.8 percentage points. These results further demonstrate its strong cross-domain generalization capability for zero-shot anomaly classification.

\noindent\textbf{Qualitative Analysis.}
Figure~\ref{fig:vis} presents qualitative comparisons of anomaly localization results on industrial and medical datasets. Compared with existing methods, SCRA-CLIP produces more accurate and complete anomaly maps, better capturing fine-grained defect regions while reducing spurious responses in normal areas. These results further demonstrate the effectiveness of our method in precise anomaly localization across diverse visual domains.

\subsection{Ablation Study}

\begin{table}[t]
\centering
\caption{Ablation study of SPCP and CGVA on MVTec-AD and VisA. I-AUROC and P-AUROC denote image-level and pixel-level AUROC (\%), respectively.}
\label{tab:ablation}
\fontsize{9}{11}\selectfont
\setlength{\tabcolsep}{1.2mm}
\begin{tabular}{cc|cc|cc}
\toprule
\multirow{2}{*}{SPCP} &
\multirow{2}{*}{CGVA} &
\multicolumn{2}{c|}{MVTec-AD} &
\multicolumn{2}{c}{VisA} \\
\cmidrule(lr){3-4}
\cmidrule(l){5-6}
& &
I-AUROC &
P-AUROC &
I-AUROC &
P-AUROC \\
\midrule
\midrule
\textcolor{black!40}{\ding{55}} & \textcolor{black!40}{\ding{55}}
& 89.0 & 90.3 & 84.1 & 94.5 \\

\ding{51} & \textcolor{black!40}{\ding{55}}
& 91.7 & 90.9 & 86.7 & 94.9 \\

\textcolor{black!40}{\ding{55}} & \ding{51}
& 90.3 & 91.7 & 85.6 & 95.6 \\

\rowcolor[HTML]{E6E6E6}
\ding{51} & \ding{51}
& \textbf{92.4} & \textbf{92.3}
& \textbf{87.2} & \textbf{96.0} \\
\bottomrule
\end{tabular}
\end{table}

\noindent\textbf{Effect of individual components.}
Table~\ref{tab:ablation} reports the ablation results of SPCP and CGVA. SPCP improves image-level AUROC by 2.7 and 2.6 percentage points on MVTec-AD and VisA, respectively, highlighting its effectiveness in enhancing anomaly-aware semantic discrimination. In comparison, CGVA yields larger gains in pixel-level AUROC than SPCP, with improvements of 1.4 and 1.1 percentage points, respectively, demonstrating its benefits for fine-grained anomaly localization. Combining both modules achieves the best performance across all metrics, with gains of 3.4/2.0 and 3.1/1.5 percentage points in image-level/pixel-level AUROC on MVTec-AD and VisA, respectively. These results demonstrate the complementary contributions of SPCP and CGVA to zero-shot anomaly detection.

\begin{table}[t]
\centering
\caption{Comparison of different prototype modeling strategies on MVTec-AD and VisA. I-AUROC and P-AUROC denote image-level and pixel-level AUROC (\%), respectively.}
\label{tab:prototype_comparison}
\renewcommand{\arraystretch}{1.2}
\resizebox{\columnwidth}{!}{
\begin{tabular}{l|cc|cc}
\toprule
\multirow{2}{*}{Prototype Modeling} &
\multicolumn{2}{c|}{MVTec-AD} &
\multicolumn{2}{c}{VisA} \\
\cmidrule(lr){2-3}
\cmidrule(l){4-5}
&
I-AUROC &
P-AUROC &
I-AUROC &
P-AUROC \\
\midrule
\midrule

K-means Prototypes
& 90.6 & 91.1 & 85.1 & 94.9 \\
Learnable Prototypes
& 90.9 & 91.5 & 85.8 & 95.3 \\
Spherical Prototypes
& 91.8 & 91.9 & 86.5 & 95.3 \\

\rowcolor[HTML]{E6E6E6}
vMF Mixture (Ours)
& \textbf{92.4} & \textbf{92.3}
& \textbf{87.2} & \textbf{96.0} \\
\bottomrule
\end{tabular}
}
\end{table}

\noindent\textbf{Effectiveness of Prototype Modeling.}
We compare SPCP with three alternative prototype strategies: K-means, learnable, and spherical prototypes. As shown in Table~\ref{tab:prototype_comparison}, performance improves progressively from K-means to learnable and spherical prototypes, with the proposed vMF mixture modeling achieving the best results across all four dataset-metric combinations. Compared with spherical prototypes, it further improves image-level/pixel-level AUROC by 0.6/0.4 and 0.7/0.7 percentage points on MVTec-AD and VisA, respectively, demonstrating the effectiveness of distribution-aware prototype modeling.

\begin{figure}
    \centering
    \includegraphics[width=\linewidth]{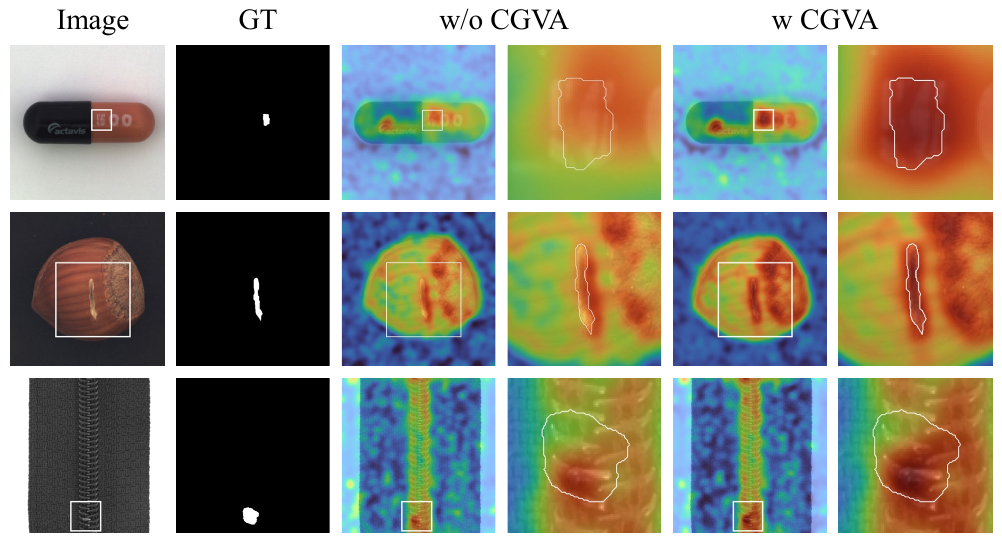}
    \caption{Qualitative comparison of anomaly localization results with and without CGVA. Zoomed-in views show the predicted anomaly responses and corresponding GT boundaries.}
    \label{fig:cgva_visualization}
\end{figure}

\noindent\textbf{Qualitative Analysis of CGVA.}
Figure~\ref{fig:cgva_visualization} presents qualitative comparisons of anomaly localization with and without CGVA. Incorporating CGVA produces more spatially focused responses within defective regions while suppressing irrelevant activations in surrounding areas. The zoomed-in views further highlight its ability to capture subtle defects, particularly small and elongated anomalies, with improved alignment to ground-truth boundaries. These results demonstrate the effectiveness of context-deviation-guided visual adaptation in refining patch representations and enhancing fine-grained anomaly localization.

\begin{table}[t]
\centering
\caption{Sensitivity analysis of the number of prototypes $K$.}
\label{tab:prototype_number}
\renewcommand{\arraystretch}{1.2}
\resizebox{\columnwidth}{!}{
\begin{tabular}{c|cc|cc}
\toprule
\multirow{2}{*}{$K$} &
\multicolumn{2}{c|}{MVTec-AD} &
\multicolumn{2}{c}{VisA} \\
\cmidrule(lr){2-3}
\cmidrule(l){4-5}
&
I-AUROC &
P-AUROC &
I-AUROC &
P-AUROC \\
\midrule
\midrule

4
& 91.3 & 91.5 & 85.8 & 95.3 \\

8
& 91.9 & 92.0 & 86.5 & 95.7 \\

\rowcolor[HTML]{E6E6E6}
16
& \textbf{92.4} & 92.3
& \textbf{87.2} & \textbf{96.0} \\

32
& 92.3 & \textbf{92.4}
& 87.0 & 95.9 \\

64
& 92.1 & 92.2 & 86.9 & 95.6 \\

\bottomrule
\end{tabular}
}
\end{table}

\noindent\textbf{Sensitivity to the Number of Prototypes.}
We evaluate the effect of the prototype number $K$ on SPCP.
As shown in Table~\ref{tab:prototype_number}, increasing $K$
from 4 to 16 improves performance on both MVTec-AD and VisA datasets, whereas
further increases bring diminishing returns and slightly
degrade most metrics. Notably, $K=16$ achieves the best
performance on three of the four metrics and remains within
0.1 percentage points of the best result on the remaining
metric, demonstrating its effectiveness as the default
configuration.

\begin{figure}
    \centering
    \includegraphics[width=\linewidth]{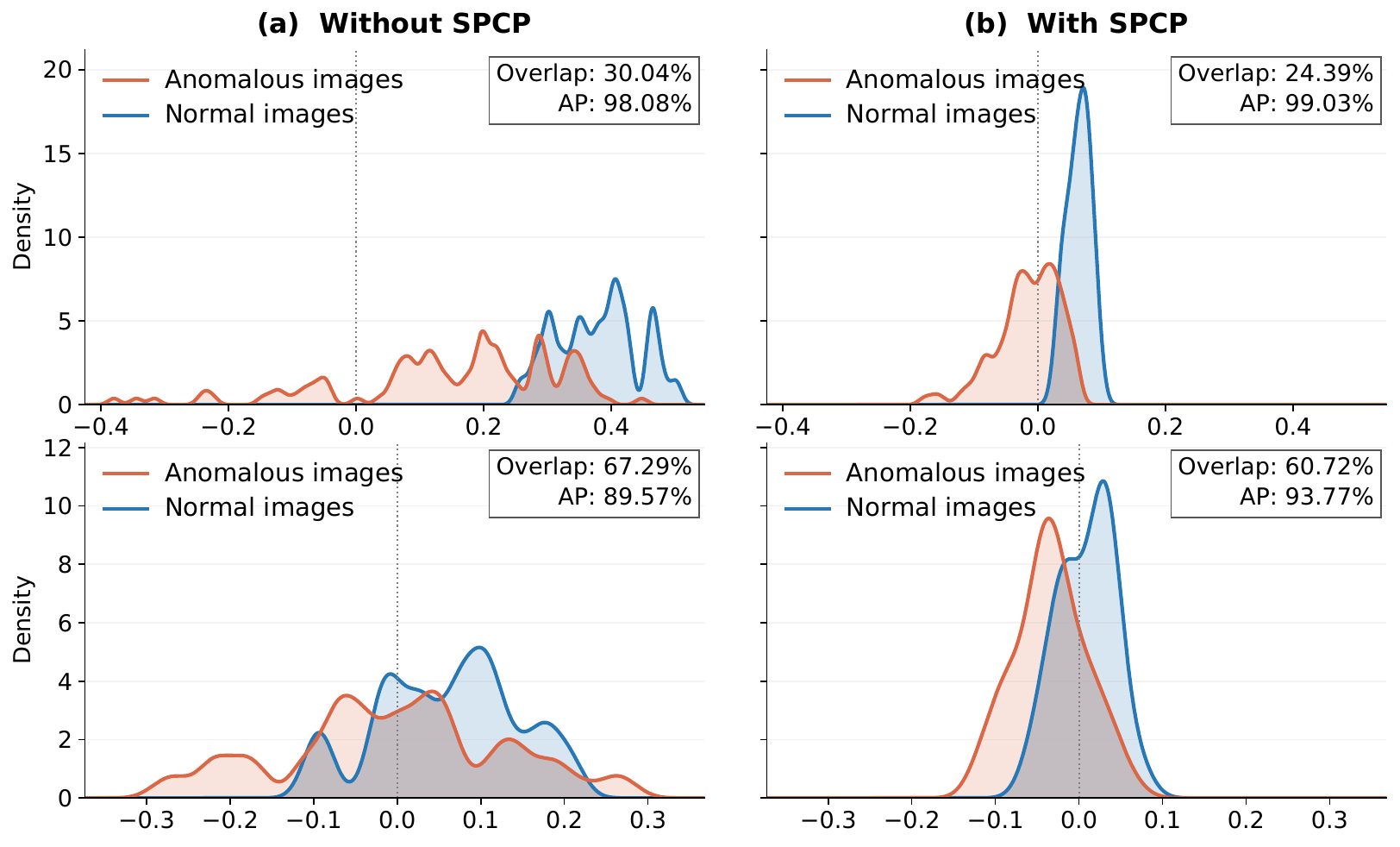}
    \caption{Image--text similarity margin distributions for normal and anomalous samples in the zipper (top row) and pill (bottom row) categories of MVTec-AD, with and without SPCP.}
    \label{fig:semantic_distribution}
\end{figure}

\noindent\textbf{Visualization of Semantic Discriminability.}
Figure~\ref{fig:semantic_distribution} illustrates the image-text similarity margin distributions of normal and anomalous samples, with and without SPCP, for the zipper (top row) and pill (bottom row) categories of MVTec-AD. SPCP reduces the distribution overlap from 30.04\% to 24.39\% on zipper and from 67.29\% to 60.72\% on pill, while simultaneously improving image-level AP from 98.08\% to 99.03\% and from 89.57\% to 93.77\%, respectively. These results demonstrate that SPCP enhances the semantic separability of normal and anomalous patterns, contributing to more accurate anomaly localization.



\begin{table}[t]
\centering
\caption{Ablation of individual auxiliary losses in SPCP.}
\label{tab:spcp_loss}
\fontsize{9}{11}\selectfont
\setlength{\tabcolsep}{1.2mm}
\begin{tabular}{cc|cc|cc}
\toprule
\multirow{2}{*}{$\mathcal{L}_{vmf}$} &
\multirow{2}{*}{$\mathcal{L}_{disc}$} &
\multicolumn{2}{c|}{MVTec-AD} &
\multicolumn{2}{c}{VisA} \\
\cmidrule(lr){3-4}
\cmidrule(l){5-6}
& &
I-AUROC &
P-AUROC &
I-AUROC &
P-AUROC \\
\midrule
\midrule

\textcolor{black!40}{\ding{55}} &
\textcolor{black!40}{\ding{55}}
& 91.2 & 91.8 & 86.0 & 95.5 \\

\textcolor{black!40}{\ding{55}} &
\ding{51}
& 91.8 & 91.9 & 86.6 & 95.6 \\

\ding{51} &
\textcolor{black!40}{\ding{55}}
& 91.5 & 92.1 & 86.3 & 95.8 \\

\rowcolor[HTML]{E6E6E6}
\ding{51} & \ding{51}
& \textbf{92.4} & \textbf{92.3}
& \textbf{87.2} & \textbf{96.0} \\

\bottomrule
\end{tabular}
\end{table}

\noindent\textbf{Effect of Auxiliary Losses in SPCP.} As shown in Table~\ref{tab:spcp_loss}, removing the vMF density-fitting
loss reduces pixel-level AUROC by 0.4 percentage points on both
MVTec-AD and VisA datasets. Removing the discriminative loss reduces image-level
AUROC by 0.9 percentage points on both datasets. The joint configuration
achieves the highest image-level and pixel-level AUROC on both benchmarks,
supporting the complementary roles of the two losses in zero-shot anomaly
classification and localization.

\section{Conclusion}

In this paper, we proposed SCRA-CLIP, a novel framework for zero-shot anomaly detection that enhances cross-modal representations by jointly improving textual semantics and visual features. Specifically, SPCP leverages vMF mixtures to model the directional distributions of normal and abnormal patch features and injects prototype-aware information into textual prompts, while CGVA captures multi-scale contextual deviations to refine local visual representations. Extensive experiments on 11 industrial and medical benchmarks demonstrate the effectiveness of SCRA-CLIP.

\section*{LLM Usage Statement}

During manuscript preparation, ChatGPT (OpenAI) was used solely for language polishing and grammatical proofreading. All suggested revisions were reviewed by the authors, who take full responsibility for the final manuscript.